\documentclass[conference]{IEEEtran}
\usepackage{times}

\usepackage[numbers]{natbib}
\usepackage{multicol}
\usepackage[bookmarks=true]{hyperref}
\usepackage{amsmath}
\usepackage{amssymb}
\usepackage{amsthm}
\usepackage{booktabs}
\usepackage{graphicx}
\usepackage{adjustbox}   
\usepackage{algorithm}
\usepackage{algorithmic}
\usepackage[capitalize,noabbrev]{cleveref}

\newif\ifpreprint
\preprinttrue
\ifpreprint\IEEEoverridecommandlockouts\fi
\ifpreprint\renewcommand{\footnoterule}{\kern-3pt\hrule width 0.4\columnwidth height 0.4pt\kern 2.6pt}\fi

\ifpreprint
\else
\fi

\newcommand{\ours}{CAL-LBLP}

\theoremstyle{plain}

\newtheorem{proposition}{Proposition}

\theoremstyle{definition}
\newtheorem{definition}{Definition}
\newtheorem{assumption}{Assumption}
\theoremstyle{remark}

\crefname{assumption}{Assumption}{Assumptions}
\crefname{remark}{Remark}{Remarks}

\newcommand{\E}{\mathbb{E}}
\newcommand{\D}{\mathcal{D}}
\newcommand{\Sset}{\mathcal{S}}
\newcommand{\Aset}{\mathcal{A}}
\newcommand{\Cset}{\mathcal{C}}
\newcommand{\M}{\mathcal{M}}
\newcommand{\sg}{\operatorname{sg}}
\newcommand{\pos}[1]{[#1]_{+}}

\begin{document}

\title{Lifted Bellman Linear Programming over\\Action Chunks for Steering VLA Policies}

\ifpreprint
\author{\authorblockN{Jongchan Park$^{1}$\thanks{$^{1}$NAVER LABS, Seongnam, Republic of Korea.
  Correspondence to: Jongchan Park $<$jcpark@naverlabs.com$>$.\protect\newline
  \protect\rule{0pt}{1.8em}\textit{Preprint. \today.}}}}
\else
\author{Author Names Omitted for Anonymous Review. Paper-ID [add your ID here]}
\fi

\maketitle

\begin{abstract}
Vision-language-action (VLA) models are trained by imitation learning, so they are limited by the quality of their demonstrations and cannot use the experience that they gather after deployment. A critic learned from this experience can score the action chunks that a frozen VLA samples at test time, but existing critics are trained by temporal-difference (TD) learning and rely on target networks and ensembles to keep it stable. We instead obtain the critic from a linear program. We formulate the Lifted Bellman Linear Program (LBLP) on the macro MDP whose actions are $h$-step action chunks, prove that its unique solution is the optimal $Q$ and $V$ over the chunks in the data, and solve it with the method of multipliers (\ours{}): a single $Q$ and $V$ on one vision-language backbone, with no target network, no $Q$ ensemble, and no call to any policy during training. Since a critic tends to overestimate chunks that it did not see in training, we use \emph{fence selection}, which removes the candidates above the upper Tukey fence of their values before taking the maximum. On the 18 Atomic-Seen tasks of RoboCasa365, one critic trained on $\pi_{0.5}$ rollouts raises the success rate of $\pi_{0.5}$ from 49.7\% to 59.2\%, against 50.8\% for the TD critic DEAS with the same architecture, data, and selection. Because the critic is not tied to a policy, it also scores the pooled chunks of $\pi_{0.5}$ and GR00T N1.5, a \emph{policy ensemble} with a single $Q$, which reaches 75.8\%, against 71.1\% for the better policy alone.
\end{abstract}

\IEEEpeerreviewmaketitle

\section{Introduction}
\label{sec:intro}
Vision-language-action (VLA) models now make decisions for robots across many tasks and scenes \citep{black2024pi0,black2025pi05,nvidia2025gr00tn1}. They are trained by imitation learning on expert demonstrations rather than by reinforcement learning (RL). Such a policy is limited by the quality of its demonstrations, and imitation learning cannot use most of the experience that a robot gathers once it is deployed: a failed rollout offers nothing to imitate, and a slow success only reproduces its inefficiency.

RL can learn from this experience, and recent work brings it to VLA models in three ways. The first trains the VLA directly, with policy gradients computed from its own online rollouts \citep{lu2025vlarl,li2025simplevlarl}. The second learns a critic from the experience and uses it to train the VLA, through the critic's gradient \citep{huang2025corft}, through advantage conditioning \citep{pi2025pistar06}, or through imitation of critic-improved actions \citep{mark2024parl}. The third does not train the VLA at all: it learns only a critic, which ranks the actions that the VLA samples at test time \citep{nakamoto2024steering,kim2025deas}. The critics of the second and third kind are mostly trained by temporal-difference (TD) learning \citep{huang2025corft,mark2024parl,nakamoto2024steering,kim2025deas}.

TD learning regresses the critic onto a target that is built from the critic's own prediction at the next state (bootstrapping), which causes three known difficulties in offline RL \citep{levine2020offline}. First, the target changes with every update of the critic, so the regression chases a moving target and can become unstable. Second, an error in the predicted value of the next state enters the target and is copied into the value of the current state, so errors accumulate from step to step along the horizon. Third, a target that maximizes over actions evaluates actions that the data do not contain. Nothing in the data corrects their values, so they are overestimated, and the overestimates corrupt the learned values.

V-GPS \citep{nakamoto2024steering} and DEAS \citep{kim2025deas} both mitigate these problems with a target network and an ensemble of $Q$ functions, and DEAS additionally with a distributional loss. In-sample methods avoid evaluating unseen actions during training \citep{kostrikov2021offline}, but at test time the critic still ranks sampled chunks that lie outside the data. In this paper, we mitigate these difficulties by obtaining the critic from a linear program (LP) instead of TD learning. The Lifted Bellman Linear Program (LBLP) writes Bellman optimality as linear inequalities on dataset pairs only \citep{yang2026lifted}. Solving it gives the optimal $Q$ and $V$ over the actions that the data contain.

AL-LBLP solves it with the method of multipliers, using a single $Q$, a single $V$, and no target network \citep{park2026allblp}. LBLP and AL-LBLP are defined for single actions, whereas a VLA acts in chunks: each time it is queried, it samples a sequence of future actions and executes its first $h$ steps open-loop \citep{zhao2023act,black2024pi0}. We therefore formulate LBLP on the \emph{macro MDP} whose actions are $h$-step chunks. We solve it with the method of AL-LBLP and call the resulting critic \ours{} (chunked AL-LBLP).

It addresses the three difficulties as follows. First, the critic minimizes an objective under inequality constraints instead of regressing onto a target built from its own predictions: the one-step row is differentiated through both the current and the next value, and only the $K$-step target is held fixed, so the critic needs no target network. Second, its $K$-step rows propagate values $K$ steps at a time, which is designed to reduce the error that bootstrapping accumulates along the horizon \citep{yang2026lifted}. Third, a critic tends to overestimate chunks that it did not see in training, so we remove the candidates whose value is too large with the Tukey fence, a standard test for outliers \citep{hoaglin1986resistant}, before taking the maximum. Because the program never evaluates a policy, the critic can be trained on any data, including failed episodes and the rollouts of several policies.

We evaluate on the 18 Atomic-Seen tasks of RoboCasa365 \citep{nasiriany2026robocasa365} with two frozen policies, $\pi_{0.5}$ \citep{black2025pi05} and GR00T N1.5 \citep{nvidia2025gr00tn1}. One critic, trained once on $\pi_{0.5}$ rollouts, raises the success rate of $\pi_{0.5}$ from 49.7\% to 59.2\% and that of GR00T N1.5 from 71.1\% to 73.9\%. Because the critic is not tied to a policy, it can also combine policies: we pool the candidate chunks of $\pi_{0.5}$ and GR00T N1.5 into one set and let the critic choose among them at every decision, which we call a \emph{policy ensemble}. The policy ensemble reaches 75.8\%, above either policy alone and close to an oracle that takes, for every task, the better of the two policies alone (76.1\%).
To our knowledge, this is the first method that selects among the action chunks of several distinct pretrained VLA policies with a single offline-trained critic over the pooled candidates.
Our contributions are:

\begin{itemize}
  \item \textbf{LBLP on the macro MDP and \ours{}.} We formulate LBLP over action chunks, prove that its unique solution is the optimal $Q$ and $V$ over the chunks that the data contain (\cref{prop:macro}), and solve it with the method of multipliers. To our knowledge, \ours{} is the first linear-programming-based offline RL method whose critic is defined over $h$-step action chunks.
  \item \textbf{Simple architecture, trained without the policy.} $Q$ and $V$ share one vision-language backbone, with no target network and no $Q$ ensemble, and are trained with a single loss and three scalar multipliers on the chunks in the data, successful or not, without querying a policy. One critic is trained once for all 18 tasks, and the architecture also improves a TD critic (\cref{tab:arch} and \cref{tab:variants}).
  \item \textbf{Fence selection.} Before taking the maximum, we remove the candidates whose value is unusually high among the candidates, which filters out overestimated out-of-distribution chunks. It raises the success rate over the argmax not only for our critic but also for a TD critic (\cref{tab:select} and \cref{tab:variants}).
  \item \textbf{Policy ensemble.} The same critic ranks the pooled candidates of two different VLAs, an ensemble of policies with a single $Q$.
\end{itemize}

\section{Related Work}
\label{sec:related}

\paragraph{Reinforcement learning for VLA models}
Recent work improves pretrained VLA models with RL in three ways. The first updates the VLA with policy gradients from online interaction: VLA-RL uses PPO with a learned value model and a learned reward model \citep{lu2025vlarl}, and SimpleVLA-RL uses GRPO on binary success rewards without a value function \citep{li2025simplevlarl}. The second learns a critic and uses it in the policy update. CO-RFT fine-tunes a VLA by offline RL with TD learning over action chunks and a conservative regularizer \citep{huang2025corft}, PA-RL trains the policy to imitate actions that a critic has re-ranked and improved \citep{mark2024parl}, and $\pi^{*}_{0.6}$ conditions the policy on advantages from a learned value function \citep{pi2025pistar06}. The third, to which our work belongs, keeps the VLA frozen and learns only a critic that ranks its samples at test time.

\paragraph{Critics for test-time selection}
V-GPS reranks the actions of generalist policies with a single-step $Q$ function learned by offline RL, and the same $Q$ function works with several different policies \citep{nakamoto2024steering}. DEAS learns a critic over action chunks and uses it for best-of-$N$ selection with GR00T N1.5 \citep{kim2025deas}; it is our main baseline. Both are trained by TD learning with a target network and an ensemble of $Q$ functions, and DEAS adds a distributional loss. Q-chunking and its decoupled variant run TD learning in the space of action chunks \citep{li2025qchunking,li2025dqc}. Other methods learn to score the candidates with TD objectives on subtasks \citep{singh2026seeq}, as action-preference models \citep{kwok2025robomonkey}, or by regression onto tree-search returns \citep{xie2026sva}. Our critic also scores whole chunks, with a single $Q$. To our knowledge, none of these methods obtains its critic from a linear program (\cref{sec:prelim}).

\paragraph{Overestimation of out-of-distribution candidates}
A critic tends to overestimate candidates that it did not see in its training data, and several selection methods avoid them. KDPE filters low-density trajectories of a diffusion policy with a kernel density estimate \citep{rosasco2025kdpe}. TACO trains a pseudo-count estimator on the internal features of the VLA and selects the candidate with the largest count \citep{yang2025taco}. SVA trains an ensemble of value heads and subtracts their standard deviation from the value \citep{xie2026sva}. We instead look only at the values that the critic gives to the $N$ candidates and remove the candidates whose value is far above the others. For this we use the Tukey fence, a standard test for outliers \citep{hoaglin1986resistant}. This method is simpler: it needs no density model, no ensemble, no additional training, and no access to the internals of the policy. We use it as the default in all our experiments, and both our critic and DEAS perform better with it than with the argmax (\cref{sec:exp-fence}).

\paragraph{Policy ensembles}
SWAP learns an offline-RL critic over policy identities that chooses which of several VLAs acts at each step \citep{das2026swap}. In multi-task RL, QMP lets a task's $Q$ function choose among the actions proposed by other tasks' policies when collecting data \citep{zhang2023qmp}. We instead pool the samples of different pretrained VLAs into one candidate set that a single critic ranks at test time, a policy ensemble with a single~$Q$.

\section{Preliminaries}
\label{sec:prelim}

\subsection{Markov decision processes}
We consider a discounted MDP $\M=(\Sset,\Aset,P,R,\gamma)$ with state space $\Sset$, action space $\Aset$, expected reward $R(s,a)$, discount $0\le\gamma<1$, and transition kernel $P(s',m\mid s,a)$ over the next state $s'$ and a mask $m\in\{0,1\}$ that is $0$ when the transition ends the episode. A policy $\pi$ has the value $Q^\pi(s,a)=\E[\sum_{t\ge0}\gamma^t(\prod_{u<t}m_u)\,r_t\mid s_0=s,a_0=a]$, and the optimal values satisfy the Bellman optimality equations
\begin{equation}
  \label{eq:bellman}
  \begin{aligned}
  & Q^*(s,a)=R(s,a)+\gamma\,\E_{(s',m)\sim P(\cdot\mid s,a)}\big[m\,V^*(s')\big],\\
  & V^*(s)=\max_a Q^*(s,a).
  \end{aligned}
\end{equation}

\subsection{Offline data and the in-support MDP}
In the offline setting we are given a dataset $\D$ of trajectories $(s_0,a_0,r_0,m_0,s_1,\dots)$ collected by one or several behavior policies. Following \citet{yang2026lifted}, the dataset $\D$ induces an empirical MDP $\M_\D$ with pair set $\mathcal X_\D=\{(s,a):(s,a,r,m,s')\in\D\}$ and action sets $\Aset_\D(s)=\{a:(s,a)\in\mathcal X_\D\}$, whose rewards and transitions are those observed in $\D$. We write $(Q^*_\D,V^*_\D)$ for its Bellman-optimal pair, the in-support optimal pair, so that $V^*_\D(s)=\max_{a\in\Aset_\D(s)}Q^*_\D(s,a)$ with the maximum taken over the actions the data contain at $s$.

\subsection{Macro MDP of action chunks}
An action chunk is a sequence of $h$ consecutive actions, $c=(a_0,\dots,a_{h-1})\in\Cset=\Aset^h$. As with options \citep{sutton1999options}, we treat each chunk as one action of a macro MDP $\M_h=(\Sset,\Cset,P_h,R_h,\gamma_h)$ with the discount $\gamma_h=\gamma^h$ per chunk. Starting from $s_0=s$ and executing $c$,
\begin{equation*}
  \begin{aligned}
  & R_h(s,c)=\E\Big[\sum_{t<h}\gamma^t M_{0:t}\,r_t\Big],\\
  & P_h(s',M\mid s,c)=\Pr\big(s_h=s',\,M_{0:h}=M\mid s_0=s,\ c\big),
  \end{aligned}
\end{equation*}
where $M_{t:t+k}=\prod_{u<k}m_{t+u}$ is the mask of the $k$ steps from $t$ and the macro mask $M_{0:h}$ is $0$ if the episode ends inside the chunk. Since $\gamma_h<1$, $\M_h$ is also a discounted MDP. Its optimal values $(Q^*_h,V^*_h)$ satisfy the Bellman optimality equations \cref{eq:bellman} of $\M_h$. Under deterministic dynamics, $V^*_h=V^*$ for any $h$.

The trajectories in the data define a macro dataset $\D_h$ with one window per time step $t$ of every trajectory (not only every $h$-th step). The window at $t$ is a transition of the macro MDP: the state $s_t$, the executed chunk $c_t=a_{t:t+h}$, its return $R^h_t=\sum_{k<h}\gamma^k M_{t:t+k}\,r_{t+k}$, its macro mask $M_{t:t+h}$, and the next state $s_{t+h}$. We write $\M_{\D,h}$ for its empirical macro MDP and $(Q^*_{\D,h},V^*_{\D,h})$ for the in-support optimal pair of $\M_{\D,h}$.

\subsection{Lifted Bellman Linear Program (LBLP)}
LBLP \citep{yang2026lifted} lifts the LP of Bellman optimality \citep{defarias2003linear} to the pair $(Q,V)$ through the epigraph inequality $Q(s,a)\le V(s)$, so that every constraint involves only dataset pairs. On $\M_\D$, with weights $\omega_Q,\omega_V>0$ and rollout length $K$,
\begin{equation}
  \label{eq:lblp}
  \begin{aligned}
  \min_{Q,V}\quad&\E_\D[\omega_Q Q(s,a)+\omega_V V(s)]\\
  \text{s.t.}\quad&g_{\mathrm B}=r+\gamma m\,V(s')-Q(s,a)\le0,\\
  &g_{\mathrm E}=Q(s,a)-V(s)\le0,\\
  &g_{\mathrm{KQ}}=\hat y_K-Q(s,a)\le0,\\
  &g_{\mathrm{KV}}=\hat y_K-V(s)\le0,
  \end{aligned}
\end{equation}
with one row of each family per dataset transition, where, for the transition at step $t$ of a trajectory, $\hat y_K$ is the $K$-step return along the data,
\begin{equation}
  \label{eq:kstep}
  \hat y_K(t)=\sum_{k<K}\gamma^k M_{t:t+k}\,r_{t+k}+\gamma^K M_{t:t+K}\,V(s_{t+K}).
\end{equation}
Its unique minimizer is $(Q^*_\D,V^*_\D)$. The $K$-step rows are implied by the others along the data, so they leave the solution unchanged while propagating values $K$ steps at a time \citep{yang2026lifted}.

\subsection{AL-LBLP}
AL-LBLP \citep{park2026allblp} solves LBLP with neural networks $Q_\theta$ and $V_\phi$ by the method of multipliers \citep{hestenes1969multiplier,rockafellar1973dual}. It writes each constraint family $F\in\{\mathrm B,\mathrm{KQ},\mathrm{KV},\mathrm E\}$ of \cref{eq:lblp} as $y^F\le0$, so that $y^F>0$ is a violation, and gives each family a scalar multiplier $\mu^F\ge0$, with one multiplier shared by the two $K$-step families. With a penalty $\rho>0$, the critic minimizes
\begin{equation}
  \label{eq:loss}
  \begin{aligned}
  \mathcal L_\rho=\E_\D\Big[\omega_Q Q_\theta+\omega_V V_\phi
  +\sum_F\tfrac{1}{2\rho}\big(\pos{\mu^F+\rho y^F}^2-(\mu^F)^2\big)\Big],
  \end{aligned}
\end{equation}
where the multipliers and the $K$-step target $\hat y_K$ are held fixed by a stop-gradient $\sg[\cdot]$. A violation enters the gradient with the weight $\pos{\mu^F+\rho y^F}$, which grows with the violation and is zero when $\mu^F+\rho y^F\le0$. After each primal step, each multiplier moves toward the mean of $\pos{\mu^F+\rho\,\sg[y^F]}$ over the data, the method-of-multipliers update.

\section{Method}
\label{sec:method}

We define the Lifted Bellman Linear Program (LBLP) on the macro MDP (\cref{sec:macro-lblp}) and solve it with the method of multipliers (\cref{sec:cal-lblp}). We then describe the critic architecture (\cref{sec:network}) and the best-of-$N$ selection with outlier rejection (\cref{sec:bon}). Proofs are in \cref{app:proofs}.

\subsection{Lifted Bellman Linear Program on the macro MDP}
\label{sec:macro-lblp}

We apply \cref{eq:lblp} to the empirical macro MDP $\M_{\D,h}$. Every per-step quantity is replaced by its macro counterpart: the action by the executed chunk, the reward by the chunk return, the discount by $\gamma_h$, the mask by the macro mask, and the rollout of $K$ steps by a rollout of $K_h$ chunks.

\begin{figure*}[!t]
  \centering
  \includegraphics[width=\textwidth]{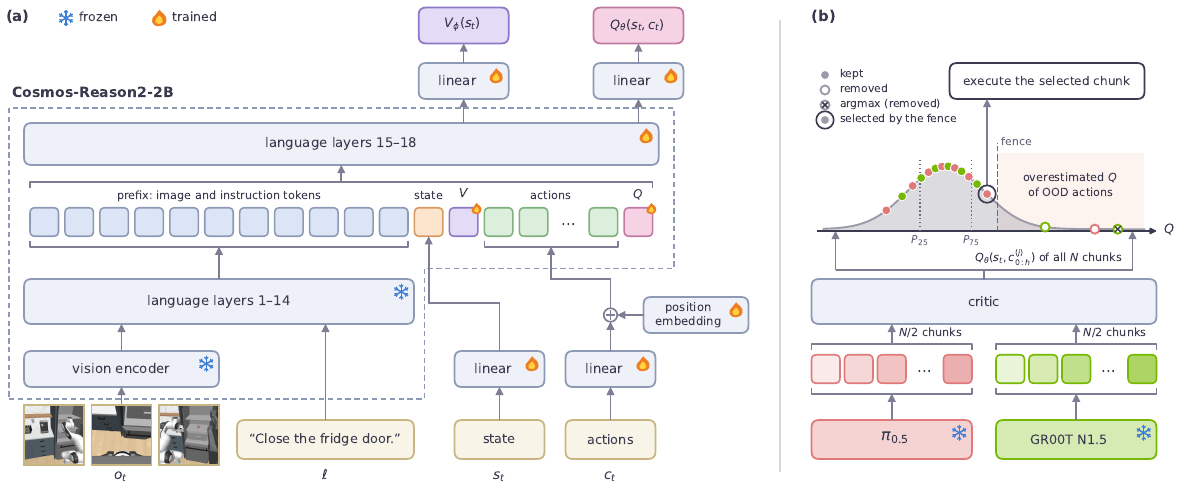}
  \vspace{-20pt}
  \caption{(a) The \ours{} critic. $V$ and $Q$ share the prefix and the state token and pass through the same fine-tuned layers 15 to 18 of Cosmos-Reason2-2B; the action and $Q$ tokens do not attend to the $V$ token, so $V$ reads [prefix, state] and $Q$ reads [prefix, state, actions]. (b) Selection at inference time: among $N$ candidate chunks, fence selection \cref{eq:bon} removes the overestimated values and the largest remaining one is executed. Candidates may come from several policies, here $\pi_{0.5}$ and GR00T N1.5 (illustrative values).}
  \label{fig:arch}
\end{figure*}

\begin{definition}[Macro LBLP]
  \label{def:macro-lblp}
  For chunk length $h$, discount $\gamma_h=\gamma^h$ per chunk, rollout length $K_h$ in chunks ($K=hK_h$ steps), and weights $\omega_Q,\omega_V>0$, the macro LBLP is
  \begin{equation}
    \label{eq:macro-lblp}
    \begin{aligned}
    \min_{Q,V}\quad&\E_{\D_h}[\omega_Q Q(s_t,c_t)+\omega_V V(s_t)]\\
    \text{s.t.}\quad&g_{\mathrm B}(t)=R^h_t+\gamma_h M_{t:t+h}\,V(s_{t+h})-Q(s_t,c_t)\le0,\\
    &g_{\mathrm E}(t)=Q(s_t,c_t)-V(s_t)\le0,\\
    &g_{\mathrm{KQ}}(t)=\hat y_{K_h}(t)-Q(s_t,c_t)\le0,\\
    &g_{\mathrm{KV}}(t)=\hat y_{K_h}(t)-V(s_t)\le0,
    \end{aligned}
  \end{equation}
  for every window $t$ of $\D_h$.
\end{definition}

Here $\hat y_{K_h}(t)$ is the return of a rollout of $K_h$ chunks from window $t$, the macro counterpart of \cref{eq:kstep},
\begin{equation}
  \label{eq:kstep-h}
  \hat y_{K_h}(t)=\sum_{j<K_h}\gamma_h^{\,j}M_{t:t+jh}\,R^h_{t+jh}+\gamma_h^{K_h}M_{t:t+hK_h}\,V(s_{t+hK_h}).
\end{equation}
It equals \cref{eq:kstep}, which is how we compute it. The rows of \cref{eq:macro-lblp} are those of \cref{eq:lblp} on $\M_{\D,h}$, with $(r,\gamma,m,s',K)$ replaced by $(R^h_t,\gamma_h,M_{t:t+h},s_{t+h},K_h)$. The rows are defined at every time step, so consecutive windows overlap and every dataset state is the start of a chunk.

\begin{assumption}[Deterministic empirical transitions]
  \label{ass:det}
  Every dataset pair $(s_t,c_t)$ has a single observed next state. This holds whenever states do not repeat across the data, as with image observations.
\end{assumption}

\begin{proposition}[Solution of the macro LBLP]
  \label{prop:macro}
  Let $h\ge1$ and $K_h\ge1$. Under \cref{ass:det}:
  \begin{enumerate}
    \item[(i)] Without the rollout rows, \cref{eq:macro-lblp} is LBLP on the discounted MDP $\M_{\D,h}$, and its unique minimizer is the in-support optimal pair $(Q^*_{\D,h},V^*_{\D,h})$.
    \item[(ii)] Every pair $(Q,V)$ that satisfies the one-step and epigraph rows satisfies the rollout rows. Adding them leaves the feasible set, and hence the minimizer, unchanged.
  \end{enumerate}
\end{proposition}

Part (i) reduces the macro program to the per-step result of \citet{yang2026lifted} with $\gamma$ replaced by $\gamma_h$. Part (ii) follows by chaining the one-step and epigraph rows of the windows $t,t+h,\dots,t+(K_h-1)h$, which are all in $\D_h$ because windows start at every step. We use $K_h=5$ chunks of $h=10$ steps.

\subsection{Chunked AL-LBLP}
\label{sec:cal-lblp}

\ours{} solves \cref{eq:macro-lblp} with the augmented Lagrangian of AL-LBLP. The networks are $Q_\theta(s,c)$, which takes the state and the chunk $c\in\mathbb R^{h\times\dim\Aset}$ as input, and $V_\phi(s)$; they share one backbone (\cref{sec:network}). The violations of window $i$ are
\begin{equation}
  \label{eq:cal-rows}
  \begin{aligned}
  &y^{\mathrm B}_i=R^h_i+\gamma_h M_{i:i+h}\,V_\phi(s_{i+h})-Q_\theta(s_i,c_i),\\
  &y^{\mathrm E}_i=Q_\theta(s_i,c_i)-V_\phi(s_i),\\
  &y^{\mathrm{KQ}}_i=\sg[\hat y_{K_h}(i)]-Q_\theta(s_i,c_i),\\
  &y^{\mathrm{KV}}_i=\sg[\hat y_{K_h}(i)]-V_\phi(s_i),
  \end{aligned}
\end{equation}
and the loss is \cref{eq:loss} with these rows. A violated row is weighted by $\pos{\mu+\rho y}$, which grows with its violation, so $\rho$ is both the step size of the dual update and the curvature of the penalty; it is the only coefficient of the relaxation. As in AL-LBLP, the two rollout families share one multiplier, so there are three scalar multipliers, $\mu^{\mathrm B}$, $\mu^{\mathrm E}$, and $\mu^{\mathrm K}$. The dual step regresses each $\mu^F=\mathrm{softplus}(\zeta^F)$ onto the batch mean of its method-of-multipliers targets $\pos{\mu^F+\rho\,\bar y^F_i}$, with $\bar y^{\mathrm K}_i=\tfrac12(y^{\mathrm{KQ}}_i+y^{\mathrm{KV}}_i)$. \Cref{alg:cal} lists one training step with plain gradient steps; in practice both steps use Adam (\cref{app:hparams}). There is no actor and no target network: a single $Q$ and a single $V$ are trained on dataset chunks only and never query a policy.

\begin{algorithm}[t]
  \caption{\ours{} critic (one training step)}
  \label{alg:cal}
  \small
  \begin{algorithmic}
    \STATE {\bfseries Input:} macro dataset $\D_h$, chunk length $h$, rollout of $K_h$ chunks, weights $\omega_Q,\omega_V$, penalty $\rho$, step sizes $\eta,\eta_\mu$
    \STATE Sample windows $\{(s_i,c_i,r_{i:i+hK_h},m_{i:i+hK_h},s_{i+h},s_{i+hK_h})\}_{i=1}^B$
    \STATE Compute $R^h_i$, $M_{i:i+h}$, $\hat y_{K_h}(i)$ with $\sg[V_\phi(s_{i+hK_h})]$ (\cref{eq:kstep-h})
    \STATE Compute the violations $y^F_i$ of \cref{eq:cal-rows}
    \STATE Primal: $(\theta,\phi)\leftarrow(\theta,\phi)-\eta\,\nabla_{\theta,\phi}\mathcal L_\rho(\theta,\phi;\sg[\mu])$ (\cref{eq:loss})
    \STATE Dual targets: $\hat\mu^F=\frac1B\sum_i\pos{\mu^F+\rho\,\bar y^F_i}$, held fixed
    \STATE Dual: $\zeta^F\leftarrow\zeta^F-\eta_\mu\nabla_{\zeta^F}(\mu^F-\hat\mu^F)^2$ for each $F$
  \end{algorithmic}
\end{algorithm}

The fixed-point property of AL-LBLP carries over because the macro rows are again affine in $(Q,V)$, with the discount and mask $\gamma m$ replaced by $\gamma_h M_{t:t+h}$: in the convex case and with one multiplier per row, the fixed points of the primal and dual updates are exactly the KKT pairs of the macro LBLP (\cref{prop:fixed}, proved in \cref{app:proofs}). The three scalar multipliers of \cref{alg:cal} approximate this per-row method, as in AL-LBLP \citep{park2026allblp}.

\subsection{Critic architecture}
\label{sec:network}

\Cref{fig:arch} shows the critic. $Q$ and $V$ share one vision-language model, Cosmos-Reason2-2B \citep{nvidia2025cosmosreason1}, whose inputs are the three camera images of the current step and the instruction. Following GR00T N1.6, which tunes the top four layers of its vision-language backbone \citep{nvidia2025gr00tn16}, we keep the vision encoder and language layers 1 to 14 frozen and fine-tune layers 15 to 18 (201\,M parameters) with a small learning rate ($10^{-5}$); the layers above 18 are not used. The newly added parts are trained with a larger learning rate ($10^{-4}$): the proprioceptive state is embedded into one token and every action of the chunk into one token by linear layers, $V$ is output by a linear head at a learned token appended to [prefix, state], and $Q$ by a second linear head at a learned token appended to [prefix, state, $a_1,\dots,a_h$]. The inputs are standardized with the statistics of the training data. Following AL-LBLP \citep{park2026allblp}, we use no target network and no ensemble, only a single linear $Q$ head and a single linear $V$ head.


\subsection{Best-of-$N$ chunk selection with outlier rejection}
\label{sec:bon}

Each time the policy $\pi$ is queried, it proposes $N$ chunks $c^{(1)},\dots,c^{(N)}\sim\pi(\cdot\mid o,\ell)$ from the observation $o$ and the instruction $\ell$. A VLA usually predicts a longer horizon $H\ge h$ and executes only its first $h$ actions. The critic scores the first $h$ actions of each proposal, $q_j=Q_\theta(s,c^{(j)}_{0:h})$ for $j=1,\dots,N$; we execute those of the selected proposal, and query the policy again at the next observation.

A critic learned by offline RL tends to overestimate out-of-distribution actions, those that the data do not contain \citep{levine2020offline}. With action chunks the problem is more severe, because a sampled chunk of $h$ actions almost never matches a chunk in the data. To avoid executing such overestimated chunks, we remove the candidates whose values are outliers before taking the maximum.

Let $P_{25}$ and $P_{75}$ be the lower and upper quartiles of the $N$ candidate values $q_1,\dots,q_N$. We discard every candidate above the upper Tukey fence \citep{hoaglin1986resistant} and execute the best of the rest,
\begin{equation}
  \label{eq:bon}
  j^\star=\operatorname*{arg\,max}_{j\,:\,q_j\le F}\,q_j,\qquad F=P_{75}+k\,(P_{75}-P_{25}),
\end{equation}
with $k=0.5$. We call this method \emph{fence selection}. It uses only the $N$ values of the current decision. It needs no ensemble and no density model, and when no value stands out it removes nothing and selects the argmax. \Cref{fig:arch}(b) illustrates the method.

\section{Experiments}
\label{sec:experiments}

We first compare \ours{} with a TD chunk critic on frozen policies, alone and as an ensemble of two (\cref{sec:exp-main}). We then analyze the architecture, the critic's training data, fence selection, the time to success, the number of candidates, how far the critic looks ahead, and the sensitivity to the penalty (\cref{sec:exp-arch,sec:exp-data,sec:exp-fence,sec:exp-time,sec:exp-n,sec:exp-lookahead,sec:exp-ablation}).

\subsection{Setup}
\label{sec:exp-setup}

\paragraph{Benchmark and policies}
We use the 18 Atomic-Seen target tasks of RoboCasa365 \citep{nasiriany2026robocasa365}, in which a mobile manipulator acts in procedurally generated kitchens, with instructions that vary with the scene. We evaluate every configuration on the same 360 scenes of the target split (20 per task) and count an episode as a success if the task succeeds within the benchmark's time limit. The base policies are frozen: a $\pi_{0.5}$ checkpoint for RoboCasa365 \citep{black2025pi05} and the GR00T N1.5 checkpoint released with the benchmark \citep{nvidia2025gr00tn1,nasiriany2026robocasa365}. Both are queried every $h=10$ steps. Alone ($N=1$), $\pi_{0.5}$ succeeds on 179 scenes (49.7\%) and GR00T N1.5 on 256 (71.1\%).

\paragraph{Critic data}
We collect 300 rollouts of the frozen $\pi_{0.5}$ per task on training scenes that are disjoint from the evaluation scenes (5{,}400 episodes, 569{,}007 windows) and keep the failed episodes as well as the successful ones. In practice, to keep the data size manageable, we start a window every 5 steps, half the chunk length, instead of at every step. A rollout runs up to twice the time limit of its task: a failure is cut at the limit, and a success after the limit is kept whole. Every critic is trained on this dataset. One critic is trained for all 18 tasks and conditioned on the instruction.

\paragraph{Training and selection}
The reward is $-1$ per step until the task succeeds and $0$ at success. A failed episode ends at its time limit with a final reward of $-1/(1-\gamma)=-1000$, the value of never succeeding, and mask 0. When a chunk runs past the end of an episode, the missing actions are filled with zero motion and the last gripper command. By default we use $h=10$, $K_h=5$ chunks, $\gamma_h=0.990$ ($\gamma=0.999$ per step), $\rho=0.05$, $\omega_Q=0.1$, $\omega_V=0.05$, 20{,}000 steps with batch size 256, learning rates $10^{-5}$ for the pretrained layers and $10^{-4}$ for the new ones, and $10^{-3}$ for the multipliers (\cref{app:hparams}). At test time the policy proposes $N=16$ chunks and we use fence selection \cref{eq:bon} with $k=0.5$. In the policy ensemble, 8 chunks come from each policy and are ranked together by the single $Q$, an ensemble of policies, not of critics. Success rates are percentages of the 360 evaluation scenes. For every critic, ours included, we report a single training seed. This configuration is used in every experiment unless a table states otherwise, and \cref{tab:hparams} lists the full setting.

\paragraph{Baseline}
DEAS \citep{kim2025deas} is a TD critic over action chunks that was used for best-of-$N$ selection with GR00T N1.5: expectile value learning, a distributional (HL-Gauss) loss, and a two-discount return. Following its official code, we train it in three settings (\cref{app:deas}): (1) its original critic architecture, an MLP on averaged frozen features of our backbone, with the optimizer setting of the DEAS paper; (2) the same architecture at our sample budget (20k steps, batch size 256); and (3) our critic architecture, with our learning rates and number of steps (\cref{tab:deas-impl} and \cref{fig:deas-arch}). DEAS without a setting number refers to setting (3). It shares our backbone and the way the backbone is fine-tuned, and differs from \ours{} in the learning algorithm (the loss, the target network, and the two $Q$ heads) and, following its official code, in the reward labelling and the discounts (\cref{app:deas}). It uses our fence selection; we also evaluate the argmax, its own selection method (\cref{tab:variants}). The tables mark setting (3) with fence selection as the default (\cref{tab:arch,tab:variants}); every other DEAS result uses it.

\subsection{Improving frozen policies and policy ensembles}
\label{sec:exp-main}

\begin{figure*}[t]
  \centering
  \includegraphics[width=\textwidth]{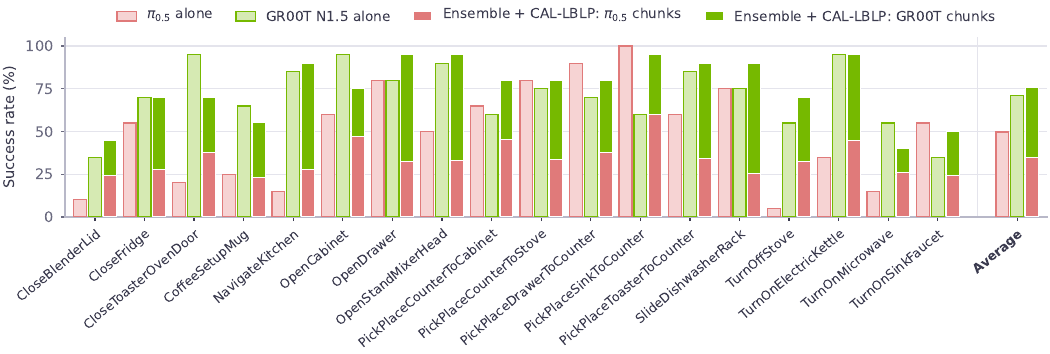}
  \vspace{-20pt}
  \caption{Success rate (\%) per task and on average. The bar of the policy ensemble with \ours{} is split in proportion to the decisions at which a $\pi_{0.5}$ or a GR00T N1.5 chunk was selected; numbers in \cref{tab:pertask}.}
  \label{fig:pertask}
\end{figure*}

\begin{table}[t]
  \centering
  \small
  \caption{Success rate (\%) on the 18 Atomic-Seen tasks of RoboCasa365, 20 scenes per task (360 scenes), best-of-16. Oracle: the better of the two policies alone on each task.}
  \label{tab:main}
  \begin{adjustbox}{max width=\columnwidth}
  \begin{tabular}{llc}
    \toprule
    Policy & Critic & Success rate (\%) \\
    \midrule
    $\pi_{0.5}$ & none & 49.7 \\
    & DEAS & 50.8 \\
    & \ours{} (ours) & \textbf{59.2} \\
    \midrule
    GR00T N1.5 & none & 71.1 \\
    & DEAS & 73.6 \\
    & \ours{} (ours) & \textbf{73.9} \\
    \midrule
    & oracle & 76.1 \\
    $\pi_{0.5}$ + GR00T N1.5 (ensemble) & DEAS & 68.1 \\
    & \ours{} (ours) & \textbf{75.8} \\
    \bottomrule
  \end{tabular}
  \end{adjustbox}
\end{table}

\Cref{tab:main} shows the main comparison on the 18 Atomic-Seen tasks of RoboCasa365, each evaluated on 20 scenes (360 in total). With \ours{}, $\pi_{0.5}$ succeeds on 59.2\% of the scenes instead of 49.7\%. DEAS with the same architecture, data, and selection method reaches 50.8\%, one point above the policy alone. With the argmax, its own selection method, it is slightly below the policy alone, and with its original architecture it is below the policy alone for $\pi_{0.5}$ in every variant (\cref{tab:variants}). For GR00T N1.5, which is already at 71.1\%, our critic improves it to 73.9\% and DEAS to 73.6\%, although both critics are trained on $\pi_{0.5}$ rollouts only. \Cref{tab:pertask,tab:family} give the results per task and per skill family.

The critic sees the images, the instruction, the state, and the chunk, not the policy's features, so it can rank the chunks of several policies together, a policy ensemble. Here it receives 8 chunks of $\pi_{0.5}$ and 8 chunks of GR00T N1.5 at every decision and is not told which policy produced which chunk. With \ours{}, the policy ensemble reaches 75.8\%, above GR00T N1.5 alone (71.1\%) and with the critic (73.9\%), and close to an oracle that takes, for every task, the better of the two policies alone (76.1\%).

\Cref{fig:pertask} shows the success rate per task. The bar of the policy ensemble is split by the share of decisions at which the critic selected a $\pi_{0.5}$ or a GR00T N1.5 chunk. Over all decisions this share is 46\% for $\pi_{0.5}$ and 54\% for GR00T N1.5, and the share of $\pi_{0.5}$ ranges from 28\% to 65\% across tasks.

On average the policy ensemble is better than either policy alone, but not on every task: it is above the better of the two policies alone on 8 of the 18 tasks, level on 3, and below on 7. On some tasks, the weaker policy is selected often and the policy ensemble still exceeds both policies. On CloseBlenderLid, $\pi_{0.5}$ alone succeeds on 10\% of the scenes and GR00T N1.5 on 35\%, yet the critic selects $\pi_{0.5}$ chunks at 54\% of the decisions and the policy ensemble reaches 45\%. On TurnOffStove, $\pi_{0.5}$ alone succeeds on 5\% and is selected at 46\% of the decisions, and the policy ensemble reaches 70\% against 55\% for GR00T N1.5. These results suggest that the critic executes GR00T N1.5 in the part where $\pi_{0.5}$ fails and $\pi_{0.5}$ in the parts it handles well, so that the two policies complement each other within an episode. A policy ensemble with a single critic can therefore exceed the better of its members, which choosing one policy per task cannot.

\subsection{Settings and architecture}
\label{sec:exp-arch}

\begin{table}[t]
  \centering
  \small
  \caption{DEAS in three settings and \ours{}, success rate (\%), all with fence selection. The settings are detailed in \cref{tab:deas-impl}.}
  \label{tab:arch}
  \setlength{\tabcolsep}{3.5pt}
  \begin{adjustbox}{max width=\columnwidth}
  \begin{tabular}{llccccc}
    \toprule
    & & & & \multicolumn{3}{c}{Policy} \\
    \cmidrule(lr){5-7}
    Critic & Architecture & Steps & Batch & $\pi_{0.5}$ & GR00T & Ensemble \\
    \midrule
    DEAS (1) & original & 30k & 64 & 36.9 & 72.2 & 50.8 \\
    DEAS (2) & original & 20k & 256 & 42.5 & 73.3 & 60.0 \\
    DEAS (3, default) & ours & 20k & 256 & 50.8 & 73.6 & 68.1 \\
    \midrule
    \ours{} (ours) & ours & 20k & 256 & \textbf{59.2} & \textbf{73.9} & \textbf{75.8} \\
    \bottomrule
  \end{tabular}
  \end{adjustbox}
\end{table}

\Cref{tab:arch} compares DEAS in three configurations with \ours{}; every critic uses our fence selection \cref{eq:bon}. \Cref{tab:deas-impl,fig:deas-arch} in \cref{app:deas} detail the DEAS settings. DEAS's original architecture, an MLP on averaged frozen features, reaches 36.9\% with $\pi_{0.5}$ at the setting of the DEAS paper (1) and 42.5\% at our sample budget (2). Replacing it by our architecture and training setup (3; \cref{fig:arch}(a) and \cref{fig:deas-arch}(b)), in which the fine-tuned language layers take the state and every action of the chunk as tokens next to the image and instruction tokens, raises it to 50.8\% with the same DEAS algorithm, and from 60.0\% to 68.1\% on the policy ensemble. Training our architecture with \ours{} instead of the DEAS algorithm raises the success rate further, to 59.2\% and 75.8\% for $\pi_{0.5}$ and the policy ensemble.

\subsection{Critic training data}
\label{sec:exp-data}

Training never queries a policy, so any data can be used. In addition to our $\pi_{0.5}$ rollouts, we collect 300 rollouts of GR00T N1.5 per task (the GR00T rollouts), and we also combine 150 rollouts of each policy per task (the $\pi_{0.5}$ + GR00T rollouts). \Cref{tab:data} compares the three sets as training data for our critic and for DEAS with the same architecture.

A critic trained on $\pi_{0.5}$ rollouts transfers to GR00T candidates without a drop in success rate: it raises GR00T N1.5 from 71.1\% to 73.9\%. The reverse does not hold: trained on GR00T rollouts alone, both critics lower the success rate of $\pi_{0.5}$ (44.7\% and 41.7\% against 49.7\% alone), although our critic still raises GR00T N1.5 to 75.0\%. The transfer is one-directional: the critic can score the chunks of any policy, but it scores those of $\pi_{0.5}$ well only when $\pi_{0.5}$ rollouts are in its data, and replacing half of the GR00T rollouts with $\pi_{0.5}$ rollouts raises $\pi_{0.5}$ from 44.7\% to 55.3\%. The training data should therefore cover the weaker, more varied policy, whose rollouts also cover the chunks of the stronger one, as the transfer from $\pi_{0.5}$ rollouts to GR00T N1.5 shows. The $\pi_{0.5}$ + GR00T rollouts give our best GR00T result (75.6\%).

Replacing all or half of the $\pi_{0.5}$ rollouts with GR00T rollouts raises DEAS above its result with $\pi_{0.5}$ rollouts alone in 4 of the 6 cases, most on the policy ensemble (from 68.1\% to 71.1\% and 74.2\%), and our critic in 2 of the 6, both for GR00T N1.5 (from 73.9\% to 75.0\% and 75.6\%). Our critic is still ahead of DEAS or level with it in 7 of the 9 settings of \cref{tab:data}; DEAS is ahead only on the policy ensemble with GR00T rollouts in the data (71.1\% against 65.0\% for the GR00T rollouts, 74.2\% against 73.9\% for the $\pi_{0.5}$ + GR00T rollouts).

\begin{table}[t]
  \centering
  \small
  \caption{Success rate (\%) by the data the critic is trained on, 300 rollouts per task in each set. Bold: the better critic per policy and data set. Oracle: the better of the two policies alone on each task.}
  \label{tab:data}
  \setlength{\tabcolsep}{4pt}
  \begin{adjustbox}{max width=\columnwidth}
  \begin{tabular}{llccc}
    \toprule
    & & \multicolumn{3}{c}{Rollouts for critic training} \\
    \cmidrule(lr){3-5}
    Policy & Critic & $\pi_{0.5}$ (default) & GR00T & $\pi_{0.5}$ + GR00T \\
    \midrule
    $\pi_{0.5}$ & none & 49.7 & 49.7 & 49.7 \\
     & DEAS & 50.8 & 41.7 & 52.8 \\
    & \ours{} (ours) & \textbf{59.2} & \textbf{44.7} & \textbf{55.3} \\
    \midrule
    GR00T N1.5 & none & 71.1 & 71.1 & 71.1 \\
     & DEAS & 73.6 & \textbf{75.0} & 71.4 \\
    & \ours{} (ours) & \textbf{73.9} & \textbf{75.0} & \textbf{75.6} \\
    \midrule
    & oracle & 76.1 & 76.1 & 76.1 \\
    Ensemble & DEAS & 68.1 & \textbf{71.1} & \textbf{74.2} \\
    & \ours{} (ours) & \textbf{75.8} & 65.0 & 73.9 \\
    \bottomrule
  \end{tabular}
  \end{adjustbox}
\end{table}

\begin{table}[t]
  \centering
  \small
  \caption{Fence selection vs.\ argmax, success rate (\%). Argmax: the candidate with the largest value. Fence: \cref{eq:bon} with factor $k$.}
  \label{tab:select}
  \setlength{\tabcolsep}{4pt}
  \begin{adjustbox}{max width=\columnwidth}
  \begin{tabular}{lccc}
    \toprule
    & \multicolumn{3}{c}{Policy} \\
    \cmidrule(lr){2-4}
    Selection & $\pi_{0.5}$ & GR00T & Ensemble \\
    \midrule
    Fence, $k=0.25$ & 58.1 & 71.7 & 72.5 \\
    Fence, $k=0.5$ (ours) & \textbf{59.2} & \textbf{73.9} & \textbf{75.8} \\
    Fence, $k=1$ & 57.5 & 71.9 & 67.8 \\
    Argmax & 51.9 & 71.4 & 64.7 \\
    \bottomrule
  \end{tabular}
  \end{adjustbox}
\end{table}

\subsection{Fence selection vs.\ argmax}
\label{sec:exp-fence}

\begin{table}[t]
  \centering
  \small
  \caption{Value rank of the chunk executed by fence selection among the 16 candidates during evaluation with $\pi_{0.5}$ (\% of decisions). The 1st is the argmax. No executed chunk ranks below 5th.}
  \label{tab:rank}
  \setlength{\tabcolsep}{4pt}
  \begin{adjustbox}{max width=\columnwidth}
  \begin{tabular}{lcccccc}
    \toprule
    Critic & 1st (argmax) & 2nd & 3rd & 4th & 5th & Total \\
    \midrule
    DEAS & 16.9 & 26.8 & \textbf{29.4} & 22.1 & 4.9 & 100 \\
    \ours{} (ours) & 18.8 & 27.2 & \textbf{29.1} & 20.6 & 4.3 & 100 \\
    \bottomrule
  \end{tabular}
  \end{adjustbox}
\end{table}

\Cref{tab:select} compares the argmax with fence selection. Executing the argmax of our critic gives 51.9\% with $\pi_{0.5}$, barely above the policy alone (49.7\%), but our fence selection raises it to 59.2\%. Fence selection is above the argmax for every critic and every candidate set, by 1.6 to 21.1 points (\cref{tab:variants}). For $\pi_{0.5}$ the fence factor matters little (58.1\%, 59.2\%, and 57.5\% for $k=0.25$, $0.5$, and $1$), but the policy ensemble is sensitive to it (72.5\%, 75.8\%, and 67.8\%). A larger factor removes fewer candidates, and as $k\to\infty$ the method becomes the argmax; already at $k=1$ the success rate is below that at $k=0.5$ for both policies and for the policy ensemble. This supports the explanation in \cref{sec:bon}: the largest of $N$ values is often an extrapolation error, for a critic from a linear program and for a TD critic alike. By removing the few values that stand out, the critic can select better chunks than the argmax does. Furthermore, when no value stands out, the fence removes nothing and selects the same chunk as the argmax.

To see which candidate fence selection executes, we record the 16 values at every decision while evaluating $\pi_{0.5}$ with fence selection on 5 scenes per task (90 episodes, about 3{,}900 decisions per critic). \Cref{tab:rank} gives the rank of the executed chunk among the 16 values: the argmax in 19\% of the decisions with \ours{} and in 17\% with DEAS, and the second to fourth largest value in most of the others. No executed chunk ranks below 5th, and the two critics give nearly the same distribution.

\subsection{Time to success}
\label{sec:exp-time}

\begin{figure*}[t]
  \centering
  \includegraphics[width=\textwidth]{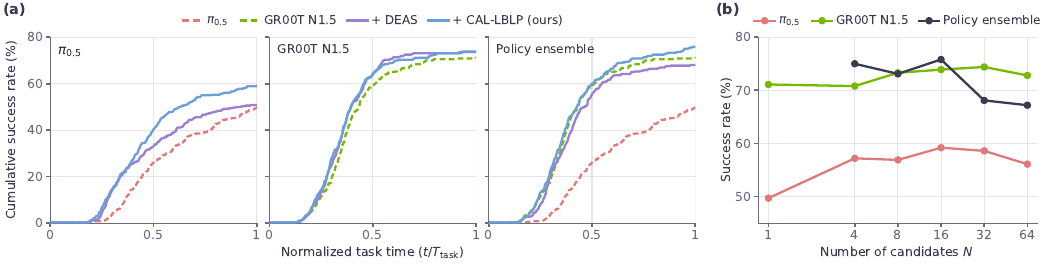}
  \vspace{-20pt}
  \caption{(a) Cumulative success rate over normalized task time $t/T_{\text{task}}$, best-of-16: $t$ is the time to success of an episode and $T_{\text{task}}$ the time limit of its task (450 to 1050 steps), pooled over the 18 tasks and 360 scenes. Task time is time in the episode, not compute time. Red and green, dashed: $\pi_{0.5}$ and GR00T N1.5 alone; purple and blue: with DEAS and with \ours{}. The ensemble panel shows both policies alone. The value at $t/T_{\text{task}}=1$ is the success rate of \cref{tab:main}. (b) Success rate (\%) over the number of candidates $N$ (log scale) with \ours{} and fence selection; $N=1$ is the policy alone, and the policy ensemble takes $N/2$ candidates from each policy.}
  \label{fig:time}
\end{figure*}

\Cref{fig:time}(a) shows the cumulative success rate over the normalized task time $t/T_{\text{task}}$, the time to success divided by the time limit of each task, pooled over the 18 tasks. With the reward of $-1$ per step (\cref{sec:exp-setup}), $Q$ is the negative discounted time to success, so \ours{} prefers the chunks that reach success soonest. \Cref{fig:time}(a) shows that with \ours{} every policy, and the policy ensemble, finishes its tasks both sooner and more often: its curve rises earlier and ends higher. With $\pi_{0.5}$, by half the time limit it has solved 40.6\% of the scenes, against 26.1\% for the policy alone and 33.1\% with DEAS, but for GR00T N1.5 the three curves are close. For the policy ensemble the curve of \ours{} stays above that of DEAS, and the gap widens toward the time limit. \Cref{tab:steps} in \cref{app:pertask} gives the median and mean episode length.

\subsection{Number of candidates}
\label{sec:exp-n}

\Cref{fig:time}(b) varies the number of candidates $N$ from 1 to 64 for the three candidate sets. With $\pi_{0.5}$, four candidates already give most of the gain: 57.2\% at $N=4$ and 59.2\% at $N=16$, against 49.7\% for the policy alone, and more candidates do not help further (56.1\% at $N=64$). GR00T N1.5 stays between 70.8\% and 74.4\% for every $N$, against 71.1\% alone. The policy ensemble is best at $N=16$ with 75.8\%, but falls to 68.1\% at $N=32$ and 67.2\% at $N=64$, below GR00T N1.5 alone. More candidates are therefore not better, and we use $N=16$ throughout.

\subsection{How far to look ahead}
\label{sec:exp-lookahead}
\begin{table}[t]
  \centering
  \small
  \caption{Discount per chunk $\gamma_h$ (rows) and rollout length $K_h$ in chunks (columns), success rate (\%) with $\pi_{0.5}$. --: not evaluated.}
  \label{tab:abl-gamma}
  \begin{adjustbox}{max width=\columnwidth}
    \begin{tabular}{lccc}
      \toprule
      $\gamma_h$ & $K_h=2$ & $K_h=5$ (default) & $K_h=10$ \\
      \midrule
      0.980 & 55.3 & 57.5 & -- \\
      0.990 (default) & 57.5 & 59.2 & 56.7 \\
      0.995 & -- & 56.9 & \textbf{60.6} \\
      \bottomrule
    \end{tabular}
  \end{adjustbox}
\end{table}

\Cref{tab:abl-gamma} varies the discount per chunk $\gamma_h$ and the rollout length $K_h$ of \ours{}, trained on the $\pi_{0.5}$ rollouts with all other settings at their defaults and evaluated with $\pi_{0.5}$. The success rate rises when $\gamma_h$ and $K_h$ grow together, that is, as the critic looks further ahead. Our default, $\gamma_h=0.990$ and $K_h=5$, follows the long-horizon setting of AL-LBLP on OGBench \citep{park2026allblp}, $\gamma=0.999$ per step and a rollout length of $K=hK_h=50$ steps, without tuning.

\subsection{Sensitivity}
\label{sec:exp-ablation}
\Cref{tab:abl-rho} varies the penalty $\rho$ of \ours{} in the same setup as \cref{tab:abl-gamma}. With every value of $\rho$, the critic raises $\pi_{0.5}$ above the policy alone (49.7\%), and unless $\rho$ is large the success rate stays robustly high: from 54.4\% to 59.2\% for $\rho$ between 0.01 and 0.5, against 51.4\% at $\rho=1$. The critic has no target-update rate and no ensemble size to tune.

\begin{table}[t]
  \centering
  \small
  \caption{Penalty $\rho$, success rate (\%) with $\pi_{0.5}$.}
  \label{tab:abl-rho}
  \begin{adjustbox}{max width=\columnwidth}
    \begin{tabular}{lccccc}
      \toprule
      $\rho$ & 0.01 & 0.05 (default) & 0.1 & 0.5 & 1 \\
      \midrule
      Success rate (\%) & 54.4 & \textbf{59.2} & 57.8 & 54.4 & 51.4 \\
      \bottomrule
    \end{tabular}
  \end{adjustbox}
\end{table}

\section{Conclusion}
\label{sec:conclusion}
We define the Lifted Bellman Linear Program on the macro MDP of action chunks, and show that its unique solution is the optimal $Q$ and $V$ over the chunks in the data. On top of this, we propose \ours{}, which solves the program with the method of multipliers to obtain a critic for frozen VLA policies: one network on one vision-language backbone with no target network, no ensemble, and no access to a policy, trained on action chunks from the dataset only. Combined with fence selection, which rejects outlying values, it improves $\pi_{0.5}$ on the 18 Atomic-Seen tasks of RoboCasa365 by about 10 points, against 1 point for a TD chunk critic with the same backbone trained on the same rollouts, and shortens the time to success. We also propose a policy ensemble, in which the critic scores the pooled candidates of two policies at every decision; it reaches 75.8\%, close to an oracle that takes, for every task, the better of the two policies alone (76.1\%).

This work has three limitations. First, all experiments are in simulation, on the 18 Atomic-Seen tasks of a single benchmark; we have not evaluated real robots or long-horizon composite tasks. Second, the policy ensemble combines only two policies, with a fixed split of 8 candidates from each. Third, steering costs computation at test time: at every decision the policy samples $N$ chunks and the critic, on a 2B vision-language backbone, scores all of them, which adds latency over the policy alone. In future work, we will extend the critic to real robots and to ensembles of more policies with other splits of the candidates. We will also train the policy with the critic, for instance toward the chunks that the critic prefers, so that the improved policy can act alone at test time, without sampling several candidates or running the critic.

\bibliographystyle{plainnat}
\bibliography{references}


\clearpage
\appendices

\section{Proofs}
\label{app:proofs}

\subsection{Notation}
In the tabular case the primal variable $x$ collects the values of $Q$ on the pairs $(s_t,c_t)$ of $\D_h$ and of $V$ on the dataset states. With linear features it collects the weight vectors of $Q$ and $V$. The objective of \cref{eq:macro-lblp} is linear, $f(x)=w^\top x$ with $w>0$, and every row of \cref{def:macro-lblp} is affine, $y_j(x)=b_j-a_j^\top x$, with $y_j(x)>0$ a violation. Masks are fixed data, so $\gamma_hM_{t:t+h}$ and $\gamma_h^{K_h}M_{t:t+hK_h}$ are constants. We write the penalty term of \cref{eq:loss} for one row as $\psi_\rho(y;\mu)=\frac{1}{2\rho}(\pos{\mu+\rho y}^2-\mu^2)$. For fixed $\mu\ge0$ it is convex and $C^1$ in $y$ with $\partial_y\psi_\rho(y;\mu)=\pos{\mu+\rho y}$: it equals $\mu y+\frac{\rho}{2}y^2$ where $\mu+\rho y>0$ and the constant $-\mu^2/(2\rho)$ elsewhere, and the two pieces and their derivatives agree on the boundary.

\subsection{Proof of \cref{prop:macro}}

\emph{(i)} Under \cref{ass:det}, $\M_{\D,h}$ is a finite MDP whose states are the dataset states, whose actions at $s_t$ are the chunks $c_t$ observed there, whose rewards are $R^h_t$, and whose transitions go to $s_{t+h}$ with mask $M_{t:t+h}$; a next state that the data never leave is absorbing with value zero, as in \citet{yang2026lifted}. Its discount is $\gamma_h\in[0,1)$. The one-step row $g_{\mathrm B}$ is the LBLP one-step row of this MDP, $r+\gamma' m V(s')-Q(s,a)$ with $(r,\gamma',m,a,s')=(R^h_t,\gamma_h,M_{t:t+h},c_t,s_{t+h})$, and $g_{\mathrm E}$ is its epigraph row. The program without rollout rows is therefore LBLP on $\M_{\D,h}$, and by Proposition~4.3 of \citet{yang2026lifted} its unique minimizer is the in-support optimal pair of $\M_{\D,h}$, which is $(Q^*_{\D,h},V^*_{\D,h})$.

\emph{(ii)} Let $(Q,V)$ satisfy every one-step and epigraph row, fix a window $t$ whose rollout row is defined, and write $t_j=t+jh$ for $j=0,\dots,K_h$. Since windows start at every step of the trajectory, the windows $t_0,\dots,t_{K_h-1}$ are all in $\D_h$, and their chunks $c_{t_j}$ are consecutive pieces of the same trajectory. Combining the epigraph row at $t_{j}$ with the one-step row at $t_j$ gives, for every $j<K_h$,
\begin{equation}
  \label{eq:chain}
  V(s_{t_j})\ge Q(s_{t_j},c_{t_j})\ge R^h_{t_j}+\gamma_hM_{t_j:t_{j+1}}V(s_{t_{j+1}}).
\end{equation}
Unrolling \cref{eq:chain} from $j=0$ to $K_h-1$, with $\prod_{l<j}M_{t_l:t_{l+1}}=M_{t:t+jh}$,
\begin{equation*}
  Q(s_t,c_t)\ge\sum_{j<K_h}\gamma_h^{\,j}M_{t:t+jh}\,R^h_{t_j}+\gamma_h^{K_h}M_{t:t+hK_h}\,V(s_{t_{K_h}})=\hat y_{K_h}(t),
\end{equation*}
the return \cref{eq:kstep-h}, so $g_{\mathrm{KQ}}(t)\le0$, and $V(s_t)\ge Q(s_t,c_t)$ gives $g_{\mathrm{KV}}(t)\le0$. Every feasible point of the program without rollout rows satisfies the rollout rows, so the two feasible sets coincide, and so do the minimizers. \qed

\subsection{Fixed points of the updates}

\Cref{sec:cal-lblp} uses that, with one multiplier per row and the update $\mu_j\leftarrow\pos{\mu_j+\rho\,y_j}$, the fixed points of the primal and dual updates are the KKT pairs of the macro LBLP; the scalar multipliers of \cref{alg:cal} approximate this per-row method \citep{park2026allblp}. We state and prove this here.

\begin{proposition}[Fixed points]
  \label{prop:fixed}
  Let the macro LBLP be a convex instance ($Q$ and $V$ tabular on $\M_{\D,h}$ or linear in fixed features) with one multiplier per row, and let $\rho>0$. A pair $((Q,V),\mu)$ is a fixed point of the primal minimization of $\mathcal L_\rho(\cdot;\mu)$ and of the dual update if and only if it is a KKT pair of \cref{eq:macro-lblp}. In the tabular case, under \cref{ass:det}, every fixed point has $(Q,V)=(Q^*_{\D,h},V^*_{\D,h})$.
\end{proposition}

\emph{Proof.} The macro program is a linear program with affine rows, and $\mathcal L_\rho(x;\mu)=w^\top x+\sum_j\psi_\rho(y_j(x);\mu_j)$ is convex and $C^1$ in $x$ with
\begin{equation*}
  \nabla_x\mathcal L_\rho(x;\mu)=w-\sum_j\pos{\mu_j+\rho\,y_j(x)}\,a_j ,
\end{equation*}
because $\psi_\rho(\cdot;\mu)$ is convex and $C^1$ with derivative $\pos{\mu+\rho y}$. A KKT pair is $(x^*,\mu^*)$ with $y_j(x^*)\le0$, $\mu^*_j\ge0$, $\mu^*_jy_j(x^*)=0$, and $w=\sum_j\mu^*_ja_j$; for a linear program with a finite optimum, $x^*$ is optimal if and only if such a $\mu^*$ exists, with no constraint qualification since the rows are affine \citep{rockafellar1970convex}.

\emph{Dual fixed point.} $\mu_j=\pos{\mu_j+\rho y_j(x)}$ holds if and only if $y_j(x)\le0$, $\mu_j\ge0$, and $\mu_jy_j(x)=0$. Indeed, if $\mu_j>0$ the positive part is attained by its argument, so $y_j(x)=0$; if $\mu_j=0$ then $\pos{\rho y_j(x)}=0$, so $y_j(x)\le0$. The converse is immediate. In either case $\pos{\mu_j+\rho y_j(x)}=\mu_j$.

\emph{Primal minimizer.} Since $\mathcal L_\rho(\cdot;\mu)$ is convex and differentiable, $x$ minimizes it if and only if $\nabla_x\mathcal L_\rho(x;\mu)=0$. Under the dual fixed point the effective multipliers equal $\mu$, so this is $w=\sum_j\mu_ja_j$, the stationarity condition of the KKT system.

Together, the two fixed-point conditions are the KKT conditions, for every $\rho>0$. In the tabular case the primal part of a KKT pair is the unique minimizer of the program, which by \cref{prop:macro} is $(Q^*_{\D,h},V^*_{\D,h})$ under \cref{ass:det}. \qed

\section{Per-task results}
\label{app:pertask}

This appendix details the results of \cref{sec:exp-main,sec:exp-time}. \Cref{tab:pertask} gives the success rate of every task for the three candidate sets (the averages are those of \cref{tab:main}), \cref{tab:family} groups the tasks into skill families, and \cref{tab:steps} gives the episode lengths behind \cref{fig:time}.

\begin{table*}[h]
  \centering
  \small
  \caption{Success rate (\%) per task (20 scenes each). Bold: the best entry of each policy group. In parentheses: the percentage of decisions at which a $\pi_{0.5}$ chunk is selected.}
  \label{tab:pertask}
  \setlength{\tabcolsep}{4pt}
  \begin{adjustbox}{max width=\textwidth}
  \begin{tabular}{lcccccccc}
    \toprule
    & \multicolumn{3}{c}{$\pi_{0.5}$} & \multicolumn{3}{c}{GR00T N1.5} & \multicolumn{2}{c}{Ensemble ($8+8$)} \\
    \cmidrule(lr){2-4}\cmidrule(lr){5-7}\cmidrule(lr){8-9}
    Task & alone & + DEAS & + \ours{} & alone & + DEAS & + \ours{} & + DEAS & + \ours{} \\
    \midrule
    CloseBlenderLid & 10.0 & 15.0 & \textbf{20.0} & 35.0 & 45.0 & \textbf{55.0} & 25.0 (45\%) & \textbf{45.0} (54\%) \\
    CloseFridge & 55.0 & 60.0 & \textbf{80.0} & 70.0 & 65.0 & \textbf{75.0} & \textbf{80.0} (39\%) & 70.0 (40\%) \\
    CloseToasterOvenDoor & 20.0 & 0.0 & \textbf{40.0} & 95.0 & 90.0 & \textbf{100.0} & 60.0 (47\%) & \textbf{70.0} (54\%) \\
    CoffeeSetupMug & 25.0 & \textbf{45.0} & 35.0 & 65.0 & \textbf{80.0} & 70.0 & \textbf{55.0} (47\%) & \textbf{55.0} (42\%) \\
    NavigateKitchen & \textbf{15.0} & 10.0 & 10.0 & \textbf{85.0} & \textbf{85.0} & 80.0 & 75.0 (40\%) & \textbf{90.0} (31\%) \\
    OpenCabinet & 60.0 & 70.0 & \textbf{80.0} & \textbf{95.0} & \textbf{95.0} & \textbf{95.0} & \textbf{75.0} (56\%) & \textbf{75.0} (63\%) \\
    OpenDrawer & 80.0 & \textbf{90.0} & 85.0 & 80.0 & 80.0 & \textbf{85.0} & 80.0 (60\%) & \textbf{95.0} (34\%) \\
    OpenStandMixerHead & 50.0 & 65.0 & \textbf{70.0} & 90.0 & \textbf{95.0} & \textbf{95.0} & \textbf{95.0} (51\%) & \textbf{95.0} (35\%) \\
    PickPlaceCounterToCabinet & 65.0 & 65.0 & \textbf{85.0} & 60.0 & \textbf{70.0} & 65.0 & 65.0 (39\%) & \textbf{80.0} (57\%) \\
    PickPlaceCounterToStove & 80.0 & 70.0 & \textbf{90.0} & 75.0 & 65.0 & \textbf{80.0} & \textbf{90.0} (47\%) & 80.0 (42\%) \\
    PickPlaceDrawerToCounter & \textbf{90.0} & 80.0 & 70.0 & \textbf{70.0} & 65.0 & 60.0 & 70.0 (46\%) & \textbf{80.0} (47\%) \\
    PickPlaceSinkToCounter & \textbf{100.0} & 90.0 & 85.0 & 60.0 & \textbf{70.0} & 65.0 & 85.0 (50\%) & \textbf{95.0} (63\%) \\
    PickPlaceToasterToCounter & 60.0 & \textbf{80.0} & \textbf{80.0} & \textbf{85.0} & 70.0 & 70.0 & 75.0 (35\%) & \textbf{90.0} (38\%) \\
    SlideDishwasherRack & 75.0 & 70.0 & \textbf{80.0} & 75.0 & \textbf{85.0} & \textbf{85.0} & 85.0 (62\%) & \textbf{90.0} (28\%) \\
    TurnOffStove & 5.0 & 10.0 & \textbf{25.0} & 55.0 & \textbf{60.0} & 55.0 & 65.0 (64\%) & \textbf{70.0} (46\%) \\
    TurnOnElectricKettle & 35.0 & 45.0 & \textbf{50.0} & \textbf{95.0} & 85.0 & \textbf{95.0} & 80.0 (57\%) & \textbf{95.0} (47\%) \\
    TurnOnMicrowave & 15.0 & \textbf{30.0} & 25.0 & 55.0 & \textbf{60.0} & 55.0 & 25.0 (65\%) & \textbf{40.0} (65\%) \\
    TurnOnSinkFaucet & \textbf{55.0} & 20.0 & \textbf{55.0} & 35.0 & \textbf{60.0} & 45.0 & 40.0 (64\%) & \textbf{50.0} (48\%) \\
    \midrule
    All 18 tasks & 49.7 & 50.8 & \textbf{59.2} & 71.1 & 73.6 & \textbf{73.9} & 68.1 (51\%) & \textbf{75.8} (46\%) \\
    \bottomrule
  \end{tabular}
  \end{adjustbox}
\end{table*}

\begin{table}[h]
  \centering
  \small
  \caption{Success rate (\%) by skill family (scenes per family in parentheses).}
  \label{tab:family}
  \setlength{\tabcolsep}{4pt}
  \begin{adjustbox}{max width=\columnwidth}
  \begin{tabular}{lccccc}
    \toprule
    & \multicolumn{2}{c}{$\pi_{0.5}$} & \multicolumn{2}{c}{GR00T N1.5} & Ensemble \\
    \cmidrule(lr){2-3}\cmidrule(lr){4-5}\cmidrule(lr){6-6}
    Skill family & alone & + ours & alone & + ours & + ours \\
    \midrule
    Open / close (140) & 50.0 & 65.0 & 77.1 & \textbf{84.3} & 77.1 \\
    Pick and place (120) & 70.0 & 74.2 & 69.2 & 68.3 & \textbf{80.0} \\
    Turn on / off (80) & 27.5 & 38.8 & 60.0 & 62.5 & \textbf{63.8} \\
    Navigation (20) & 15.0 & 10.0 & 85.0 & 80.0 & \textbf{90.0} \\
    \midrule
    All (360) & 49.7 & 59.2 & 71.1 & 73.9 & \textbf{75.8} \\
    \bottomrule
  \end{tabular}
  \end{adjustbox}
\end{table}

\begin{table}[h]
  \centering
  \small
  \caption{Episode length in steps, median and mean over all 360 episodes, a failure counted at the time limit of its task (lower is better; the counterpart of \cref{fig:time}).}
  \label{tab:steps}
  \begin{adjustbox}{max width=\columnwidth}
  \begin{tabular}{llcc}
    \toprule
    Policy & Critic & Median & Mean \\
    \midrule
    $\pi_{0.5}$ & none & 450 & 497 \\
    & DEAS & 450 & 463 \\
    & \ours{} (ours) & \textbf{438} & \textbf{431} \\
    \midrule
    GR00T N1.5 & none & 294 & 376 \\
    & DEAS & 272 & \textbf{354} \\
    & \ours{} (ours) & \textbf{264} & \textbf{354} \\
    \midrule
    $\pi_{0.5}$ + GR00T N1.5 (ensemble) & DEAS & 310 & 388 \\
    & \ours{} (ours) & \textbf{281} & \textbf{360} \\
    \bottomrule
  \end{tabular}
  \end{adjustbox}
\end{table}

\section{Ablations}
\label{app:ablation}

\begin{table}[h]
  \centering
  \small
  \caption{Critic and selection variants, success rate (\%). Argmax: the candidate with the largest value, which is also DEAS's own selection method. DEAS settings (1) to (3): \cref{tab:deas-impl}. Default and ours: the DEAS and \ours{} configurations used in all other experiments.}
  \label{tab:variants}
  \setlength{\tabcolsep}{4pt}
  \begin{adjustbox}{max width=\columnwidth}
  \begin{tabular}{lccc}
    \toprule
    Critic and selection & $\pi_{0.5}$ & GR00T N1.5 & Ensemble \\
    \midrule
    None ($N=1$) & 49.7 & 71.1 & -- \\
    \midrule
    DEAS (1), argmax & 23.1 & 66.7 & 36.7 \\
    DEAS (2), argmax & 28.6 & 70.6 & 38.9 \\
    DEAS (3), argmax & 49.2 & 66.7 & 61.7 \\
    \ours{}, argmax & 51.9 & 71.4 & 64.7 \\
    \midrule
    DEAS (1), fence & 36.9 & 72.2 & 50.8 \\
    DEAS (2), fence & 42.5 & 73.3 & 60.0 \\
    DEAS (3), fence (default) & 50.8 & 73.6 & 68.1 \\
    \ours{}, fence (ours) & \textbf{59.2} & \textbf{73.9} & \textbf{75.8} \\
    \bottomrule
  \end{tabular}
  \end{adjustbox}
\end{table}

\section{DEAS baseline}
\label{app:deas}

\begin{figure*}[!t]
  \centering
  \includegraphics[width=\textwidth]{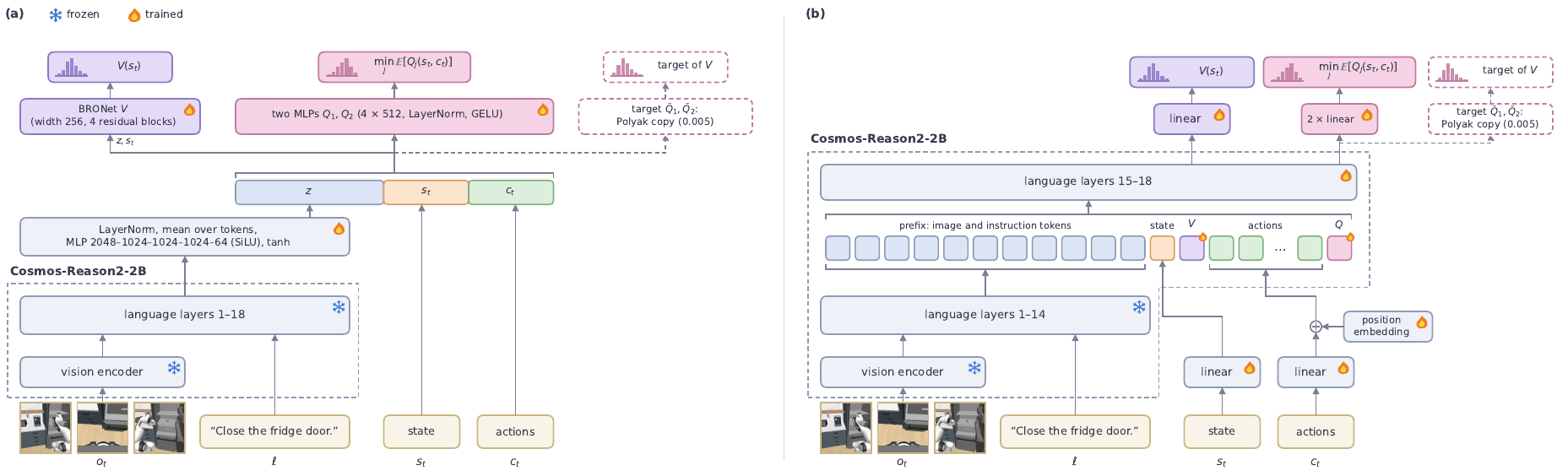}
  \vspace{-10pt}
  \caption{The two DEAS critic architectures: (a) the original one, settings (1) and (2); (b) ours, setting (3).}
  \label{fig:deas-arch}
\end{figure*}

We follow the official implementation of DEAS \citep{kim2025deas}. The last 15 steps of a successful episode are labelled as success and the reward is shifted to $\{-1,0\}$. We keep the discounts of the official setting, which treat the steps inside a chunk and the chunks differently: the return of a chunk of $h=10$ steps uses the discount $0.9$ per step, and the value after the chunk is discounted by $0.99$ per step, that is $0.99^{10}\approx0.904$ per chunk, against $\gamma=0.999$ per step ($\gamma_h=0.990$ per chunk) for \ours{}. $V$ is trained by expectile regression (0.7) toward the target critic, and the two $Q$ heads by a distributional HL-Gauss cross-entropy loss with 101 atoms on $[-100,0]$, bootstrapping from the online $V$. At test time the candidate value is the minimum over the two heads of the expected value. \Cref{tab:deas-impl,fig:deas-arch} give the three settings. Our implementation of the original architecture omits DEAS's self-attention module over the backbone tokens, a trained part of GR00T N1.5 that our backbone has no counterpart for.

\begin{table*}[!t]
  \centering
  \small
  \caption{The three DEAS settings. ``As (1)'': the same as setting (1).}
  \label{tab:deas-impl}
  \begin{tabular}{p{2.6cm}p{5.0cm}p{2.4cm}p{5.4cm}}
    \toprule
    & (1) Original architecture, paper setting & (2) Original architecture, our budget & (3) Our architecture \\
    \midrule
    Backbone & Cosmos-Reason2-2B, layers 1--18 frozen & as (1) & layers 1--14 frozen, layers 15--18 fine-tuned \\
    Backbone output & LayerNorm, mean over tokens, MLP encoder (2048--1024--1024--1024--64, SiLU), tanh & as (1) & last hidden state of a learned $V$ or $Q$ token \\
    State and chunk & concatenated with the encoded feature & as (1) & one token for the state and one per action, input to layers 15--18 \\
    $V$ & BRONet (width 256, 4 residual blocks) on [feature, state] & as (1) & linear head \\
    $Q$ & two MLPs ($4\times512$, LayerNorm, GELU) on [feature, state, chunk] & as (1) & two linear heads \\
    Outputs & 101 bins on $[-100,0]$; $Q$: minimum over the heads of the expectation & as (1) & as (1) \\
    Target critic & Polyak copy (0.005) of the $Q$ MLPs & as (1) & Polyak copy (0.005) of the two linear $Q$ heads, applied to the online network's $Q$ token \\
    Standardized inputs & state & as (1) & state and actions \\
    Optimizer & Adam, $3\cdot10^{-4}$, constant, gradient clip 1.0 & as (1) & AdamW (weight decay $10^{-5}$), $10^{-5}$ / $10^{-4}$, cosine, 5\% warm-up, no clip \\
    Steps, batch size & 30k, 64 & 20k, 256 & 20k, 256 \\
    \bottomrule
  \end{tabular}
\end{table*}

\section{Hyperparameters}
\label{app:hparams}

\paragraph{Discount and rollout length}
In the idealized iteration of \citet{yang2026lifted} the error contracts at the discount of the one-step rows, and at the discount of the rollout on the rows where the rollout term is active. On $\M_{\D,h}$ the one-step rows contract at $\gamma_h$ per chunk, and the rollout rows by $\gamma_h^{K_h}$ every $K_h$ chunks. We use $\gamma_h=0.990$ ($\gamma=0.999$ per step of a chunk of $h=10$) and $K_h=5$, so that $\gamma_h^{K_h}=0.951$ matches the long-horizon setting of AL-LBLP \citep{park2026allblp}.

\begin{table*}[h]
  \centering
  \caption{Default setting of \ours{} in all experiments.}
  \label{tab:hparams}
  \begin{adjustbox}{max width=\textwidth}
  \begin{tabular}{ll}
    \toprule
    Chunk length $h$, rollout length $K_h$ & 10 steps, 5 chunks \\
    Discount per chunk $\gamma_h$ & 0.990 ($\gamma=0.999$ per step, $\gamma_h^{K_h}=0.951$) \\
    Penalty $\rho$ & 0.05 \\
    Objective weights $\omega_Q,\omega_V$ & 0.1, 0.05 \\
    Multipliers & three scalars ($\mu^{\mathrm B},\mu^{\mathrm E},\mu^{\mathrm K}$), initial value $10^{-3}$ \\
    Reward & $-1$ per step, $0$ at success, $-1000$ at the end of a failure \\
    Backbone & Cosmos-Reason2-2B, vision encoder and layers 1--14 frozen, \\
    & layers 15--18 trained (201\,M parameters) \\
    Inputs & 3 cameras, instruction, state (16), chunk ($10\times12$) \\
    Optimizer & Adam, weight decay $10^{-5}$, cosine schedule, \\
    & 5\% warm-up \\
    Learning rates & $10^{-5}$ (pretrained layers), $10^{-4}$ (new layers), \\
    & $10^{-3}$ (multipliers) \\
    Steps, batch size & 20{,}000, 256 \\
    Candidates $N$, fence factor $k$ & 16, 0.5 \\
    \bottomrule
  \end{tabular}
  \end{adjustbox}
\end{table*}

\end{document}